\documentclass[journal]{IEEEtran}
\ifCLASSINFOpdf
\else
\fi

\usepackage{amssymb}
\usepackage{url}

\usepackage{graphicx} 
\usepackage{subfigure}
\usepackage{multirow}
\usepackage{hhline}
\usepackage{array}
\usepackage{algorithm}
\usepackage{algorithmic}
\usepackage[margin=1in]{geometry}
\usepackage{amsmath}
\usepackage{amssymb}
\usepackage{epstopdf}

\usepackage{color}

\def\colorshenb{\color{black}}

\urldef{\mailsa}\path|{alfred.hofmann, ursula.barth, ingrid.haas, frank.holzwarth,|
	\urldef{\mailsb}\path|anna.kramer, leonie.kunz, christine.reiss, nicole.sator,|
	\urldef{\mailsc}\path|erika.siebert-cole, peter.strasser, lncs}@springer.com|

\begin{document}
%
\title{Deep Low-Shot Learning for Biological Image Classification and Visualization from Limited Training Samples}
%
%
%

\author{Lei Cai, Zhengyang Wang, Rob Kulathinal, Sudhir Kumar, Shuiwang Ji ~\IEEEmembership{Senior~Member,~IEEE}
\thanks{Lei Cai is with the School of Electrical Engineering and Computer Science, Washington State University, Pullman, 
WA 99164. E-mail: lei.cai@wsu.edu.}
\thanks{Zhengyang Wang and Shuiwang Ji are with the Department of Computer Science and Engineering, Texas A\&M University,
	College Station, TX 77843. E-mail: \{zhengyang.wang, sji\}@tamu.edu.}
\thanks{Rob Kulathinal and Sudhir Kumar are with the Department of Biology, Temple University, Philadelphia, PA 19122. E-mail: \{robkulathinalu, s.kumar\}@temple.edu.}}

%
%

\markboth{IEEE Transactions on Neural Networks and Learning Systems}%
{Cai \MakeLowercase{\textit{et al.}}: }
%



\maketitle

\begin{abstract}
Predictive modeling is useful but very challenging in biological
image analysis due to the high cost of obtaining and labeling
training data. For example, in the study of gene interaction and
regulation in \textit{Drosophila} embryogenesis, the analysis is
most biologically meaningful when \textit{in situ} hybridization
(ISH) gene expression pattern images from the same developmental
stage are compared. However, labeling training data with precise
stages is very time-consuming even for developmental biologists.
Thus, a critical challenge is how to build accurate computational
models for precise developmental stage classification from limited
training samples. In addition, identification and visualization of
developmental landmarks are required to enable biologists to
interpret prediction results and calibrate models. To address these
challenges, we propose a deep two-step low-shot learning framework
to accurately classify ISH images using limited training images.
Specifically, to enable accurate model training on limited training
samples, we formulate the task as a deep low-shot learning problem
and develop a novel two-step learning approach, including data-level
learning and feature-level learning. We use a deep residual network
as our base model and achieve improved performance in the precise
stage prediction task of ISH images. Furthermore, the deep model can
be interpreted by computing saliency maps, which consist of
pixel-wise contributions of an image to its prediction result.
In our task, saliency maps are used to assist the identification and
visualization of developmental landmarks. Our experimental results
show that the proposed model can not only make accurate predictions,
but also yield biologically meaningful interpretations. We
anticipate our methods to be easily generalizable to other
biological image classification tasks with small training datasets.
Our open-source code is available at
\url{https://github.com/divelab/lsl-fly}.
\end{abstract}


%
\IEEEpeerreviewmaketitle

\section{Introduction}

\IEEEPARstart{I}{n} biological image analysis tasks, annotation of training images
usually requires specific domain knowledge and thus is commonly
performed by expert biologists. This manual practice and the
incurred cost limit the number of labeled training samples.
Therefore, a recurring theme in biological image analysis is how to
enable efficient and accurate model training from limited labeled
training samples \cite{Zeng:BMC15}. For example, comparative
analysis of \textit{in situ} hybridization~(ISH) gene expression
pattern images is a key step in studying gene interactions and
regulations in \textit{Drosophila} embryogenesis. Advances in
imaging technologies have enabled biologists to collect an
increasing number of \textit{Drosophila} embryonic ISH
images~\cite{cardona2012current,walter2010visualization,peng2007automatic}.
Currently, more than one hundred thousand \textit{Drosophila} ISH
images are available for investigating the functions and
interconnections of
genes~\cite{lecuyer2007global,tomancak2002systematic,daugman1980two,levine2005gene}.
The analysis of gene expression patterns is most biologically
meaningful when images from the same developmental stage are
compared~\cite{KumarBest,Frise:MSB,PengRECOMB}. Moreover,
identification and visualization of developmental landmarks in each
stage are required to enable biologists to interpret prediction
results.

\begin{figure*}[t]
	\centering
	\includegraphics[width=\textwidth]{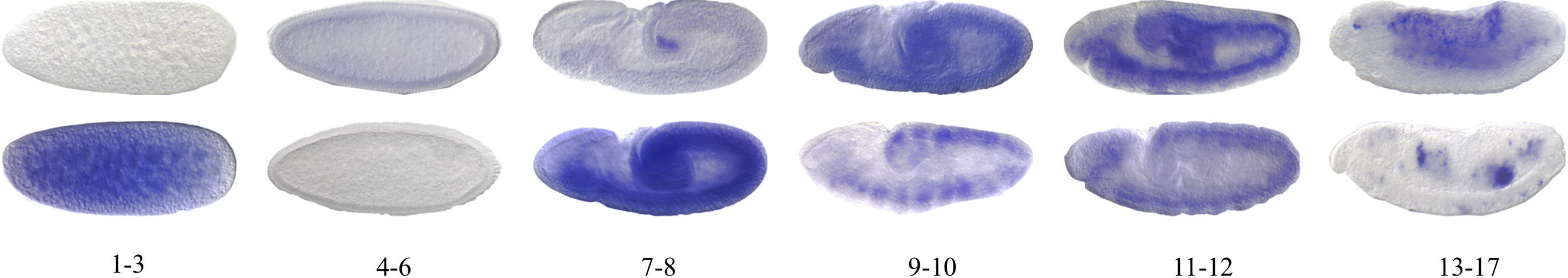}
	\caption{Examples of \textit{Drosophila} ISH images in different stage ranges. The images are annotated with stage range labels 1-3, 4-6, 7-8, 9-10, 11-12 and 13-17~\cite{tomancak2002systematic,tomancak2007global}.
	}
	\label{fig:embryo}
\end{figure*}

However, ISH images obtained from high-throughput experiments are
commonly annotated with stage range labels, as illustrated by Figure~\ref{fig:embryo}.
since determining the precise stage for each ISH image is
very hard and time-consuming even for expert biologists. Currently,
most of the comparative analysis is limited to stage range levels
due to the lack of accurate and cost-effective methods for assigning
precise stage labels to images. In~\cite{yuan2013automated}, a
small training set is manually labeled with precise stages by
developmental biologists, and a computational pipeline is proposed
to train classifiers from this small training set.
Specifically, pre-defined Gabor filters are used to compute features from
ISH images~\cite{daugman1980two,field1987relations} and linear
classifiers are employed to predict the precise stages of ISH images. In that
work, no attempt is made to explicitly consider and account for the
key bottleneck of limited training samples. Due to the small size of
training set and the methods used, the predictive performance
is relatively poor in that work. In order for the results to be
usable by routine biological studies, an accurate computational
model that can be trained effectively from limited training samples is
highly desirable~\cite{ji2008automated,ye2006classification,cai2012joint,Hartenstein:book}.

We notice that the Gabor filter features used in~\cite{yuan2013automated} are computed by convolving a set of
pre-defined filters to input images. Although the pre-defined Gabor
filters have been widely used, the filter weights are pre-computed and extract features
that do not adapt to each task and data specifically. In order to
enable automated data-driven and task-related feature extraction, we propose to
employ deep convolutional neural networks (DCNNs) for this task~\cite{he2016deep}. However, DCNNs require a large number of
training samples to estimate the filter weights while most biological
applications, including the precise stage prediction task of ISH images that we
are considering, have only very limited training samples. Thus,
straightforward use of CNNs inevitably results in poor performance.

In this work, we propose a deep two-step low-shot learning framework
that consists of a set of novel computational techniques to enable
effective deep model training on a small number of annotated
training samples. Although we focus on ISH image classification and
visualization in this work, the proposed approach is generic and can
be easily generalizable to other biological image analysis tasks.
Specifically, we formulate the original classification task with
limited training samples as a deep low-shot learning
problem~\cite{fei2006one,lake2011one,wang2018low} and propose a deep
two-step low-shot learning approach that consists of data-level
learning and feature-level learning. The first data-level learning
step leverages extra data to facilitate the training. In the precise
stage prediction task of ISH images, we can access a large number of
extra ISH images with stage range labels, which cannot be directly
used to train a precise stage prediction model. However, these extra
ISH images are obtained from the same experimental pipeline and
follow the same distribution as images in the original small
training set. Therefore, they can be exploited to guide the
training. The second step of our low-shot learning method is
feature-level learning. We propose a regularization method which
forces a high similarity of features extracted from different
samples in the same class. It results in a high-quality feature
space, enabling the extracted features to generalize well on unseen
data. In order to achieve the regularization for classification, we
generate reference sets that contain representative samples of each
class. The proposed feature-level learning step is to fine-tune the
model by adding a similarity loss based on cosine similarities
between features extracted from the training sample and reference
samples. The proposed deep two-step low-shot learning framework
enables us to effectively train deep learning models with high
classification accuracy. In addition, we explore the interpretation
of deep learning models through saliency
maps~\cite{simonyan2013deep}, which indicate pixel-wise
contributions of input images to the prediction results. In the
precise stage prediction task of ISH images, we generate masked
genomewide expression maps
(GEMs)~\cite{konikoff2012comparison,kumar2011flyexpress} based on
saliency maps. The masked GEMs provide biologically meaningful
visualizations that help identifying and visualizing developmental
landmarks.

\section{A Deep Two-Step Low-Shot Learning Framework}

In this section, we introduce our deep two-step low-shot learning framework. We start with the motivation, advantages and challenges of applying deep learning on biological image classification in Section~\ref{sec:challenge}. To overcome the challenges, we formulate the task as a deep low-shot learning problem and propose the deep two-step low-shot learning approach to address it. The two steps, namely data-level learning and feature-level learning, are described in Sections~\ref{sec:first_step} and~\ref{sec:second_step}, respectively. In Section~\ref{sec:application}, we apply the proposed method with a deep residual network on the precise stage prediction task of ISH images. We also introduce the interpretation method for deep learning models and how to use it to generate biologically meaningful visualizations for our task in Section~\ref{sec:interpret}.

\subsection{Deep Learning for Biological Image Classification}\label{sec:challenge}

\begin{figure*}[t]
	\centering
	\includegraphics[width=\textwidth]{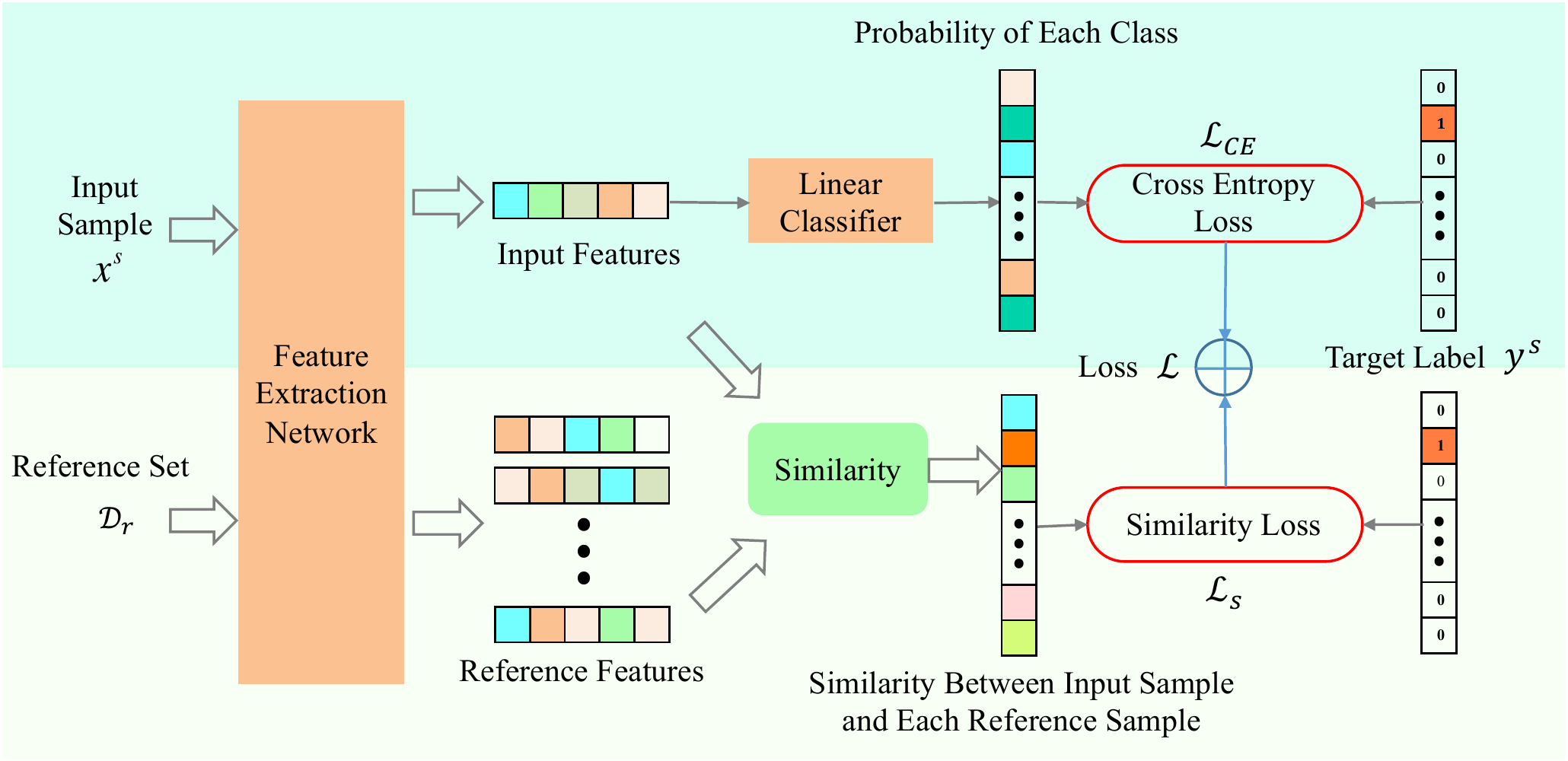}
	\caption{Illustration of feature-level deep low-shot learning. A training sample and a reference
		set are fed into the feature extraction network to generate features.
		We compute cosine similarity of features between the input sample and each
		reference sample and propose the similarity loss $\mathcal{L}_S$. Then, the similarity loss $\mathcal{L}_S$
		and the cross entropy loss ${eq:loss}$ are combined to train the network.}
	\label{fig:featurelevel}
\end{figure*}

Many traditional biological image analyses
involve a hand-crafted feature extraction step, which requires
expertise and only focuses on specific types of features. An example
is using the Gabor filter to obtain texture features for the stage
prediction task of ISH images~\cite{yuan2013automated}. While
hand-crafted features achieve acceptable performance in practice,
they have two main disadvantages. First, it is expensive and
time-consuming to design these feature extractors. There are a
variety of biological images with different tasks, and each of which
needs a re-design of feature extraction methods accordingly. Second,
the performance is limited by human's knowledge. However, more
accurate analysis is needed to improve our understanding.

Deep learning has provided a promising way to perform automatic
feature extraction in a data-driven and task-related
manner~\cite{lecun1998gradient}. Instead of relying on hand-crafted
feature extractors, deep neural networks can learn to extract
features through training on annotated data. Specifically, experts
are only required to analyze and annotate a limited amount of data
manually. Then, deep neural networks are able to determine features
that are important to the task and design corresponding feature
extractors. Based on them, the trained networks can perform the task
on unlabeled data automatically, in an efficient and accurate way.

A key challenge of applying deep learning for biological image
analysis is that the number of available annotated data is very
limited, while deep learning models typically require a large amount
of training data. It is usually cost-effective to perform coarse
annotation on large-scale data. Fine annotation with high accuracy
can only be done on a very small portion of data due to practical
limitations of resources. It hinders the wide use of deep learning.

To address the challenge, we formulate it as a deep low-shot
learning problem. In this problem, a large dataset with coarse
labels and a small dataset with fine labels are available. To be
specific, we denote the large dataset as $\mathcal{D}_t=\{(x_i^t,
y_i^t)\}_{i=1}^{n_t}$ and the small dataset as
$\mathcal{D}_s=\{(x_i^s, y_i^s)\}_{i=1}^{n_s}$, where $y_i^t\in
\{1,2,\ldots,l_t\}$ and $y_i^s\in\{1,2,\ldots,l_s\}$, respectively.
Here, $n_t$ and $n_s$ represent the number of annotated data in each
dataset and we have $n_t>>n_s$. Correspondingly, $l_t$ and $l_s$
represent the number of classes in coarse and fine annotations,
respectively. In this setting, it is reasonable to assume $l_t<l_s$.
Note that $y_i^t=1$ does not have the same meaning as $y_i^s=1$,
since coarse and fine annotations have different label space. The
data samples in both datasets, $x_i^t$ and $x_i^s$, are from the
same distribution. The task is to develop accurate deep learning
models to perform fine annotation. However, the small $n_s$ prevents
us to directly train a deep neural network on $\mathcal{D}_s$,
making it necessary to develop a method that is able to effectively
leverage $\mathcal{D}_t$. To solve this problem, we propose a method
combining data-level learning and feature-level learning, leading to
our deep two-step low-shot learning approach.

\begin{figure*}[t]
	\centering
	\includegraphics[width=\textwidth]{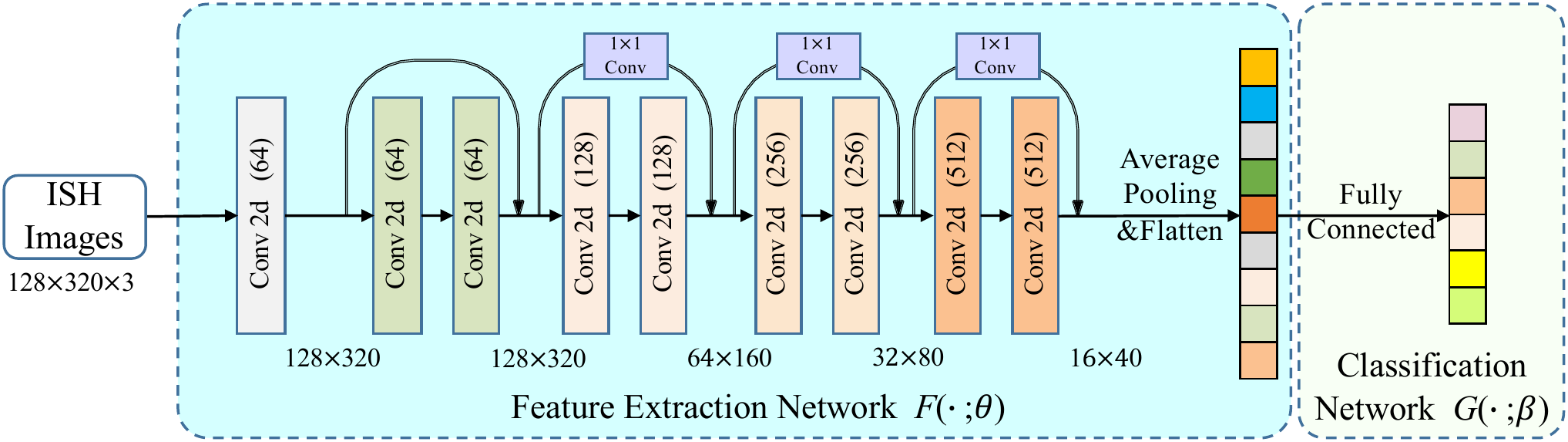}
	\caption{The architecture of our deep residual network~(ResNet) in the precise stage prediction task of ISH images. The feature extraction network $F(\cdot;\theta)$ contains four residual blocks. An average pooling with a kernel size of $4\times4$ is employed after the last residual block. As illustrated in Algorithm~\ref{alg: alg}, the same $F(\cdot;\theta)$ is employed through the whole two-step process. The classification network $G(\cdot;\beta)$ refers to $G_t(\cdot;\beta_t)$ and $G_s(\cdot;\beta_s)$ in corresponding steps.}
	\label{fig:resnet}
\end{figure*}

\subsection{Data-Level Deep Low-Shot Learning}\label{sec:first_step}

The first step of our proposed approach is data-level learning. Namely, we propose a way to leverage the large dataset $\mathcal{D}_t$ and learn from more data, even when these data have coarse labels only. To introduce our method, we first review the deep classification networks.

A deep classification network is typically composed of a feature extraction network and a classification network. The feature extraction network corresponds to the hand-crafted feature extraction step in traditional methods and the classification network performs the task based on extracted features. We focus on the feature extraction network as extracting features of high quality is the key to achieving satisfactory classification performance. Based on good features, the task can be well solved by using simple machine learning methods like nearest neighbor algorithm and linear classification methods~\cite{snell2017prototypical}. Actually, the classification networks in modern deep classification networks~\cite{he2016deep,he2016identity} are usually equivalent to multi-class logistic regression. The classification network simply serves as a guide for feature extraction network to extract task-related features from given data.

Note that, in the deep low-shot learning problem defined in Section~\ref{sec:challenge}, $x_i^t$ and $x_i^s$ are from the same distribution. In addition, $y_t$ and $y_s$ are labels of two different accuracy levels for the same classification task. With the insights above, we point out that training a deep classification network for the coarse annotation task on $\mathcal{D}_t$ is able to provide an informative and useful feature extraction network for the ultimate fine annotation task.

Therefore, our data-level learning step proposes to first train a deep classification network on $\mathcal{D}_t$, where a large amount of annotated data are available. Specifically, the deep classification network consisting of a feature extraction network $F(\cdot;\theta)$ and a classification network $G_t(\cdot;\beta_t)$ is denoted as $G_t(F(\cdot;\theta);\beta_t)$, where $\theta$ and $\beta_t$ represent the parameters in $F(\cdot)$ and $G_t(\cdot)$, respectively. Note that $G_t(\cdot;\beta_t)$ performs classification based on coarse labels, whose output layer has $l_t$ nodes. After training $G_t(F(\cdot;\theta);\beta_t)$ on $\mathcal{D}_t$, we build another deep classification network $G_s(F(\cdot;\theta);\beta_s)$ to perform fine annotation. In particular, we employ the same extraction network $F(\cdot;\theta)$ and simply change the classification network to $G_s(\cdot;\beta_s)$ whose output layer has $l_s$ nodes. Finally, we train $G_s(F(\cdot;\theta);\beta_s)$ on $\mathcal{D}_s$. In terms of training, we let the outputs of the classification network to go through a \textit{Softmax} function and compute the cross-entropy loss for back-propagation~\cite{lecun1998gradient}. Taking $G_s(F(\cdot;\theta);\beta_s)$ as an example, for one training sample $(x^s, y^s)$ in $\mathcal{D}_s$, the \textit{Softmax} function is defined as:
\begin{equation} \label{eq:softmax0}
p_i=\frac{\exp(o_i)}{\sum_{j=1}^{l_s}{\rm exp}(o_j)}, i=1,2,\ldots,l_s,
\end{equation}
where $o_i$ denote the output of the $i$-th node, \emph{i.e.},
\begin{equation} \label{eq:output}
o_i=(G_s(F(x^s;\theta);\beta_s))_i.
\end{equation}
Here, $p_i$ can be interpreted as the probability that the input image $x^s$ belongs to the $i$-th class. The cross-entropy loss can be computed by:
\begin{equation} \label{eq:loss}
\mathcal{L}_{CE}=-{\rm log}(p_{y^s}),
\end{equation}
which simply depends on the predicted probability that $x^s$ belongs to the $y^s$-th class.

The key step of data-level learning is to apply the same feature extraction network in the process. Intuitively, such an approach enables the task-related knowledge obtained from $\mathcal{D}_t$ to be transferred to $\mathcal{D}_s$. Technically, training $F(\cdot;\theta)$ on $\mathcal{D}_t$ provides a good initialization for training $G_s(F(\cdot;\theta);\beta_s)$ on $\mathcal{D}_s$, alleviating the problem of lacking enough training data for the fine annotation task.

\subsection{Feature-Level Deep Low-Shot Learning}\label{sec:second_step}

The second step of our proposed approach is feature-level learning, where we propose a regularization method to force a high similarity of features extracted from different samples in the same class. To achieve it, we select several samples from each class as reference samples and regularize the feature extraction network to extract similar features from other samples in the same class. We first describe how to select reference samples and then introduce the regularization method.

After the data-level learning step described in Section~\ref{sec:first_step}, we obtain a deep classification network $G_s(F(\cdot;\theta);\beta_s)$ that has been trained on $\mathcal{D}_s$. Using the \textit{Softmax} function in Equation~(\ref{eq:softmax0}), for each training sample $(x^s, y^s)$ in $\mathcal{D}_s$, we can compute the probability $p_{y^s}$ that $x^s$ belongs to $y^s$. We set a threshold $\tau$ and select those samples with $p_{y^s}>\tau$. The setting of $\tau$ should make sure that $p_{y^s}>p_i, \forall i\in\{1,2,\ldots,l_s\}\backslash\{y^s\}$. Specifically, for each class $c=1,2,\ldots,l_s$, we obtain a subset of $\mathcal{D}_s$ defined as $\mathcal{D}_{s,c}=\{(x_c^s, c)\ |\ p_c>p_i, \forall i\in\{1,2,\ldots,l_s\}\backslash\{c\}\ {\rm and}\ p_c>\tau\}$. Note that these subsets $\mathcal{D}_{s,c}$, $c=1,2,\ldots,l_s$ are mutually exclusive. Then, we randomly select one sample from each $\mathcal{D}_{s,c}$ to form a reference set $\mathcal{D}_r=\{(x_c^r, c)\}_{c=1}^{l_s}$ which contains exactly $l_s$ reference samples. The last step is iterated for $k$ times to form $k$ reference sets $\mathcal{D}_{r_i}$, $i=1,2,\ldots,k$.

The feature-level learning step follows the data-level learning by fine-tuning $G_s(F(\cdot;\theta);\beta_s)$ on $\mathcal{D}_s$ with the help of reference sets $\mathcal{D}_{r_i}$, $i=1,2,\ldots,k$. When fine-tuning with one training sample $(x^s, y^s)$ in $\mathcal{D}_s$, we randomly pick a reference set $\mathcal{D}_r=\{(x_c^r, c)\}_{c=1}^{l_s}$ from $\mathcal{D}_{r_i}$, $i=1,2,\ldots,k$. The feature-level learning aims to force the features extracted from input images in the same class to be similar, \emph{i.e.}, $F(x^s;\theta)$ and $F(x_{y_s}^r;\theta)$ should be similar. In opposite, for $c\in\{1,2,\ldots,l_s\}\backslash\{y^s\}$, $F(x^s;\theta)$ and $F(x_c^r;\theta)$ should be less similar. It requires a quantitative way to evaluate the similarity between features. In this work, we quantify the similarity by calculating:
\begin{eqnarray}
{\rm sim}(x^s, x_c^r) &=& {\rm cosine}(F(x^s),F(x_c^r)) \nonumber\\ 
&=&\frac{F(x^s)\cdot F(x_c^r)}{||F(x^s)||\times||F(x_c^r)||}.
\end{eqnarray}
For each $(x_c^r,c) \in \mathcal{D}_r$, we can compute the corresponding ${\rm sim}(x^s, x_c^r)$. Then, we normalize the similarities through a \textit{Softmax} function:
\begin{equation}\label{eq: sim}
{\rm SIM}(x^s, x_c^r) = \frac{\exp({\rm sim}(x^s, x_c^r))}{\sum_{i=1}^{l_s}{\rm exp}({\rm sim}(x^s,x_i^r))}, c=1,2,\ldots,l_s,
\end{equation}
where a large ${\rm SIM}(x^s, x_c^r)$ means that $F(x^s;\theta)$ and $F(x_c^r;\theta)$ are similar. At last, we propose the similarity loss:
\begin{eqnarray}
\mathcal{L}_S = &-&\ {\rm SIM}(x^s, x_{y^s}^r) \nonumber\\  &+& \frac{1}{l_s-1}\sum_{c\in\{1,2,\ldots,l_s\}\backslash\{y^s\}}{\rm SIM}(x^s, x_c^r) .
\end{eqnarray}
Minimizing the similarity loss $\mathcal{L}_S$ will result in maximizing ${\rm SIM}(x^s, x_{y^s}^r)$ while minimizing ${\rm SIM}(x^s, x_c^r)$ for $i\in\{1,2,\ldots,l_s\}\backslash\{y^s\}$. This corresponds to our purpose of forcing features extracted from input images in the same class to be similar as well as making those in different classes distinguishable. Therefore, the similarity loss $\mathcal{L}_S$ serves as a feature-level regularization method. When fine-tuning the deep classification network in the feature-level learning step, we apply a combined loss function defined as:
\begin{equation} \label{eq:loss1}
\mathcal{L} = \mathcal{L}_{CE} + \mathcal{L}_S,
\end{equation}
where $\mathcal{L}_{CE}$ is defined by Equation~(\ref{eq:loss}).

An overview of the proposed feature-level learning step is provided in Figure~\ref{fig:featurelevel}. The feature-level learning provides better features for the fine annotation task and improves $G_s(F(\cdot;\theta);\beta_s)$.

\subsection{Deep Two-Step Low-Shot Learning for the Precise Stage Prediction of ISH Images}\label{sec:application}

\begin{algorithm}[t]
	\renewcommand{\algorithmicrequire}{\textbf{Input:}}
	\renewcommand{\algorithmicensure}{\textbf{Step}}
	\setlength{\tabcolsep}{3.2pt}
	\caption{Deep Two-step Low-shot Learning for Biological Image Classification.}\label{alg: alg}
	\begin{algorithmic}
		\REQUIRE ~~\\ {\colorshenb Dataset with Coarse Labels $\mathcal{D}_t=\{(x_i^t, y_i^t)\}_{i=1}^{n_t}$ where $y_i^t\in \{1,\ldots,l_t\}$ \\
			Dataset with Fine Labels $\mathcal{D}_s=\{(x_i^s, y_i^s)\}_{i=1}^{n_s}$ where $y_i^s\in\{1,\ldots,l_s\}$} \\
		Threshold $\tau$
		\ENSURE 1 Data-Level Low-shot Learning:~~\\
		\STATE 1. Train $G_t(F(\cdot;\theta);\beta_t)$ on $\mathcal{D}_t$.
		\STATE 2. Train $G_s(F(\cdot;\theta);\beta_s)$ with the same feature extraction network $F(\cdot;\theta)$ on $\mathcal{D}_s$.
		\ENSURE 2: Feature-Level Low-shot Learning:~~\\
		\STATE 3. Construct reference sets $\mathcal{D}_{r_i}$, $i=1,2,\ldots,k$ from $\mathcal{D}_s$ using $G_s(F(\cdot;\theta);\beta_s)$ and $\tau$.
		\STATE 4. Fine-tune $G_s(F(\cdot;\theta);\beta_s)$ using the loss function defined in Equation~(\ref{eq:loss1}).
	\end{algorithmic}
\end{algorithm}

Combining data-level learning and feature-level learning leads to our proposed deep two-step low-shot learning approach, shown in Algorithm~\ref{alg: alg}. We apply this approach to solve the precise stage prediction task of \textit{Drosophila} ISH images~\cite{yuan2013automated}. In this task, the large dataset with coarse labels $\mathcal{D}_t$ refers to the ISH images with stage range labels as illustrated in Figure~\ref{fig:embryo}. These labels only indicate that the developmental stage of an ISH image falls into one of the six ranges. Meanwhile, a small dataset with fine labels $\mathcal{D}_s$ is available, where the data are annotated with precise stages. Here, $l_t=6$ and $l_s=14$, respectively. The precise stage prediction task can be well formulated as a deep low-shot learning problem and addressed by our proposed method.

\begin{figure*}[t]
	\centering
	\includegraphics[width=\textwidth]{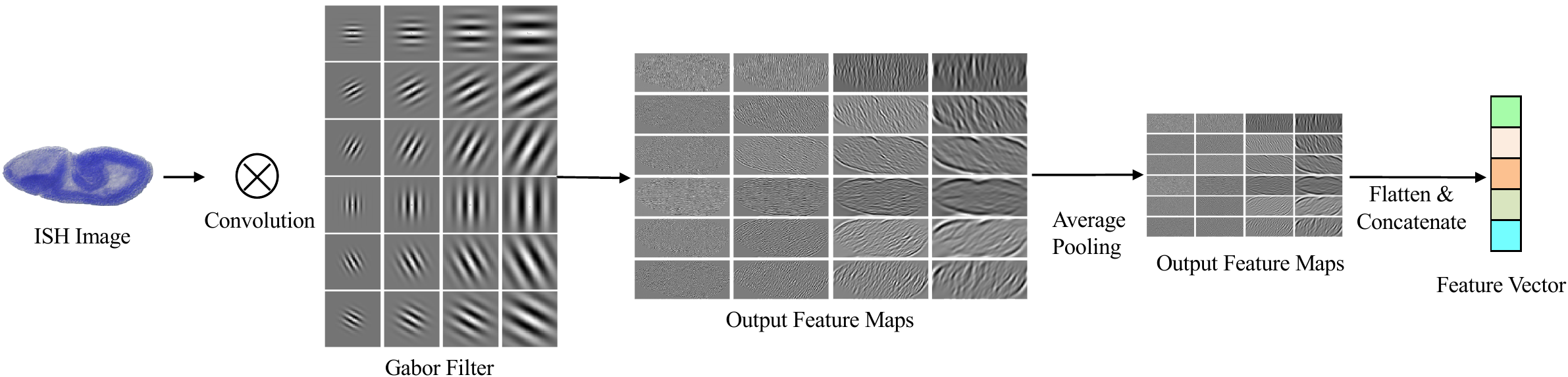}
	\caption{Illustration of precise stage prediction using Gabor filters~\cite{yuan2013automated}. In
		this approach, 24 filters are generated with four different wavelet
		scales and six different filter orientations. The average pooling
		layer with kernel size $8\times8$ is employed to reduce the spatial
		size of the feature maps and extract the invariant features. Finally, the
		extracted features are flattened and concatenated together.}
	\label{fig:gabor_pipeline}
\end{figure*}

Deep convolutional neural networks~(DCNNs)~\cite{lecun1998gradient} have achieved great success in various image tasks. Convolutional layers are proved to be capable of performing effective feature extraction. Therefore, we use DCNNs as our deep learning model for the precise stage prediction task. In particular, we apply a deep residual network (ResNet) by adopting the technique of adding residual connections proposed in~\cite{he2016deep}. Residual connections are known to facilitate the training of deep neural networks.

Figure~\ref{fig:resnet} illustrates our deep residual network for the stage prediction task. Our feature extraction network $F(\cdot;\theta)$ contains four residual blocks, each of which is composed of two convolutional layers and a residual connection. A convolutional layer contains a convolution followed by a batch normalization~\cite{ioffe2015batch} and a ReLU activation function~\cite{krizhevsky2012imagenet}. A residual connection is employed to add the inputs to the outputs for two consecutive convolutional layers and forms a residual block. The residual connection is simply an identity connection for the first residual block. For the latter three residual blocks, we set the stride to 2 for the first convolutional layer, resulting in outputs of reduced spatial sizes. In this case, we use a $1\times1$ convolutional layer in the residual connection, which changes the spatial sizes of inputs accordingly to accommodate the addition. We apply an average pooling with a kernel size of $4\times4$ to obtain the extracted features after the last residual block.
The classification networks $G_t(\cdot;\beta_t)$ and $G_s(\cdot;\beta_s)$ are designed as a single fully-connected layer with $6$ and $14$ output nodes, respectively. Our deep learning model is trained by following Algorithm~\ref{alg: alg} and outperforms the previous state-of-the-art model~\cite{yuan2013automated} significantly, as shown in Section~\ref{sec:experiment}.

\subsection{Interpretation and Visualization Using Saliency Maps}\label{sec:interpret}

Our proposed deep two-step low-shot learning approach yields deep learning models with high accuracy. We further introduce an interpretation method for deep learning models based on saliency maps. Saliency maps quantify the pixel-wise contributions of an input image to its prediction result by computing the gradient of the prediction outputs with respect to the input image~\cite{simonyan2013deep}. To be specific, to interpret our deep classification network $G_s(F(\cdot;\theta);\beta_s)$ with input image $x^s$, we compute the saliency maps for each output node:
\begin{equation}
S(x^s,i) = \frac{\partial o_i}{\partial x^s}, i=1,2,\ldots,l_s,
\end{equation}
where $o_i$ is defined in Equation~(\ref{eq:output}). Suppose the prediction result of $x^s$ is $y^s$. We pay attention to $S(x^s,y^s)$ as it tells which pixels in $x^s$ support the prediction result most.

In the precise stage prediction task of ISH images, we apply saliency maps to generate masked genomewide expression maps~(GEMs) to help identifying and visualizing developmental landmarks. GEMs provide a visualization method for precise stage prediction models and a visualization method for
developmental landmark analysis. Originally, GEMs are generated by aggregating and
normalizing the ISH images from the same stage~\cite{konikoff2012comparison,kumar2011flyexpress}.
We propose to use saliency maps as masks in generating GEMs, selecting characteristic pixels from ISH images that contribute to the prediction results. We show that the resulted masked GEMs can provide more biologically meaningful visualizations in Section~\ref{sec:experiment}.

\begin{table*}[t]
	\caption{An overview of the dataset used in this study.}
	\label{table: data}
	\begin{center}
		\begin{tabular}{|c|c|c|c|c|c|}
			\hline
			\multicolumn{1}{|c|}{Stage} &
			\begin{tabular}[c]{@{}c@{}}\# images with\\precise stage labels\end{tabular} &
			\multicolumn{1}{c|}{\begin{tabular}[c]{@{}c@{}}\# images with\\stage range labels\end{tabular}} &
			\multicolumn{1}{c|}{Stage} &
			\multicolumn{1}{c|}{\begin{tabular}[c]{@{}c@{}}\# images with\\precise stage labels\end{tabular}} &
			\multicolumn{1}{c|}{\begin{tabular}[c]{@{}c@{}}\# images with\\stage range labels\end{tabular}} \\ \hline
			1-3                         & 0                                                                                 & 4203                                                                                            & 11                         & 246                                                                                                   & \multirow{2}{*}{9248}                                                                           \\ \cline{1-5}
			4                           & 251                                                                               & \multirow{3}{*}{7344}                                                                           & 12                         & 255                                                                                                   &                                                                                                 \\ \cline{1-2} \cline{4-6}
			5                           & 274                                                                               &                                                                                                 & 13                         & 251                                                                                                   & \multirow{5}{*}{8680}                                                                           \\ \cline{1-2} \cline{4-5}
			6                           & 224                                                                               &                                                                                                 & 14                         & 252                                                                                                   &                                                                                                 \\ \cline{1-5}
			7                           & 236                                                                               & \multirow{2}{*}{3665}                                                                           & 15                         & 232                                                                                                   &                                                                                                 \\ \cline{1-2} \cline{4-5}
			8                           & 260                                                                               &                                                                                                 & 16                         & 243                                                                                                   &                                                                                                 \\ \cline{1-5}
			9                           & 248                                                                               & \multirow{2}{*}{3768}                                                                           & 17                         & 254                                                                                                   &                                                                                                 \\ \cline{1-2} \cline{4-6}
			10                          & 248                                                                               &                                                                                                 & Total                      & 3474                                                                                                  & 36908                                                                                           \\ \hline
		\end{tabular}
	\end{center}
\end{table*}

\begin{table*}[h]
	\caption{Comparison of performance between different methods in terms of precise stage prediction accuracy. The first model is our baseline~\cite{yuan2013automated}. The last three rows correspond to ResNet models trained using three different approaches: train it directly on $\mathcal{D}_s$~(Plain ResNet), train it using data-level low-shot learning only~(Data-level), and train it using both data-level and feature-level low-shot learning~(Data\&Feature-level).} \label{table: res}
	\begin{center}
		\begin{tabular}{cccccccccccccccc}
			\hline
			Model & \begin{tabular}[c]{@{}c@{}}Stage  4\end{tabular} & \begin{tabular}[c]{@{}c@{}}Stage  5\end{tabular} & \begin{tabular}[c]{@{}c@{}}Stage  6\end{tabular} & \begin{tabular}[c]{@{}c@{}}Stage  7\end{tabular} & \begin{tabular}[c]{@{}c@{}}Stage  8\end{tabular} & \begin{tabular}[c]{@{}c@{}}Stage  9\end{tabular} & \begin{tabular}[c]{@{}c@{}}Stage  10\end{tabular} & \begin{tabular}[c]{@{}c@{}}Stage  11\end{tabular}  \\
			\hline
			Gabor Filter \cite{yuan2013automated} & \textbf{84\%} & 83\% & \textbf{91\%} & 95\% & 82\% & 62\% & 64\% & 85\% \\
			Plain ResNet & 81\% & 77\% & 81\% & 90\% & 80\% & 57\% & 53\% & 83\%  \\
			Data-level & 76\% & \textbf{91\%} & 85\% & \textbf{99\%} & 84\% & 66\% & \textbf{66\%} & \textbf{86\%}  \\
			Data\&Feature-level & 81\% & 89\% & 87\% & 98\% & \textbf{87\%} & \textbf{75\%} & 64\% & 84\%  \\
			\hline
			\\
			\hline
			Model &  \begin{tabular}[c]{@{}c@{}}Stage  12\end{tabular} & \begin{tabular}[c]{@{}c@{}}Stage  13\end{tabular} & \begin{tabular}[c]{@{}c@{}}Stage  14\end{tabular} & \begin{tabular}[c]{@{}c@{}}Stage  15\end{tabular} & \begin{tabular}[c]{@{}c@{}}Stage  16\end{tabular} & \begin{tabular}[c]{@{}c@{}}Stage  17\end{tabular} & Average  \\
			\hline
			Gabor Filter \cite{yuan2013automated} & 80\% & 94\% & 74\% & 86\% & 73\% & 80\% & 80.97\%\\
			Plain ResNet & 80\% & 89\% & 77\% & 77\% & 72\% & 80\% & 77.28\% \\
			Data-level & 88\% & 92\% & \textbf{84\%} & 74\% & 71\% & 84\% & 82.40\% \\
			Data\&Feature-level & \textbf{89\%} & \textbf{94\%} & 79\% & \textbf{81\%} & \textbf{74\%} & \textbf{87\%} & \textbf{83.92\%} \\
			\hline
		\end{tabular}
	\end{center}
\end{table*}

\section{Related Work}

Currently, the state-of-the-art method~\cite{yuan2013automated} for the precise stage prediction
task of ISH images is to extract features using pre-defined convolutional filters and train a linear
classifier based on the extracted features.
Texture features of different scales and orientations can be
computed with different filters. The filter weights are constructed
by using log Gabor filters~\cite{daugman1980two,field1987relations}.
Given a set of pre-defined wavelet scales and orientations, we can generate
corresponding Gabor filters to extract the specific features by convolving on input images.
The precise stage prediction model in~\cite{yuan2013automated} based on Gabor filter features is
illustrated in Figure~\ref{fig:gabor_pipeline}.

The Gabor filter features focus on subtle textures which enable the
classification model to distinguish ISH images from different
stages. However, there are still some limitations for Gabor filter
features as Gabor filters are hand-crafted and cannot learn
features automatically for different tasks and datasets. Moreover, Gabor
filters only extract low-level texture features. Extracting features hierarchically from low-level to high-level is
appealing for improving the precise stage prediction accuracy.

\begin{figure*}[!ht]
	\centering
	\includegraphics[width=\textwidth]{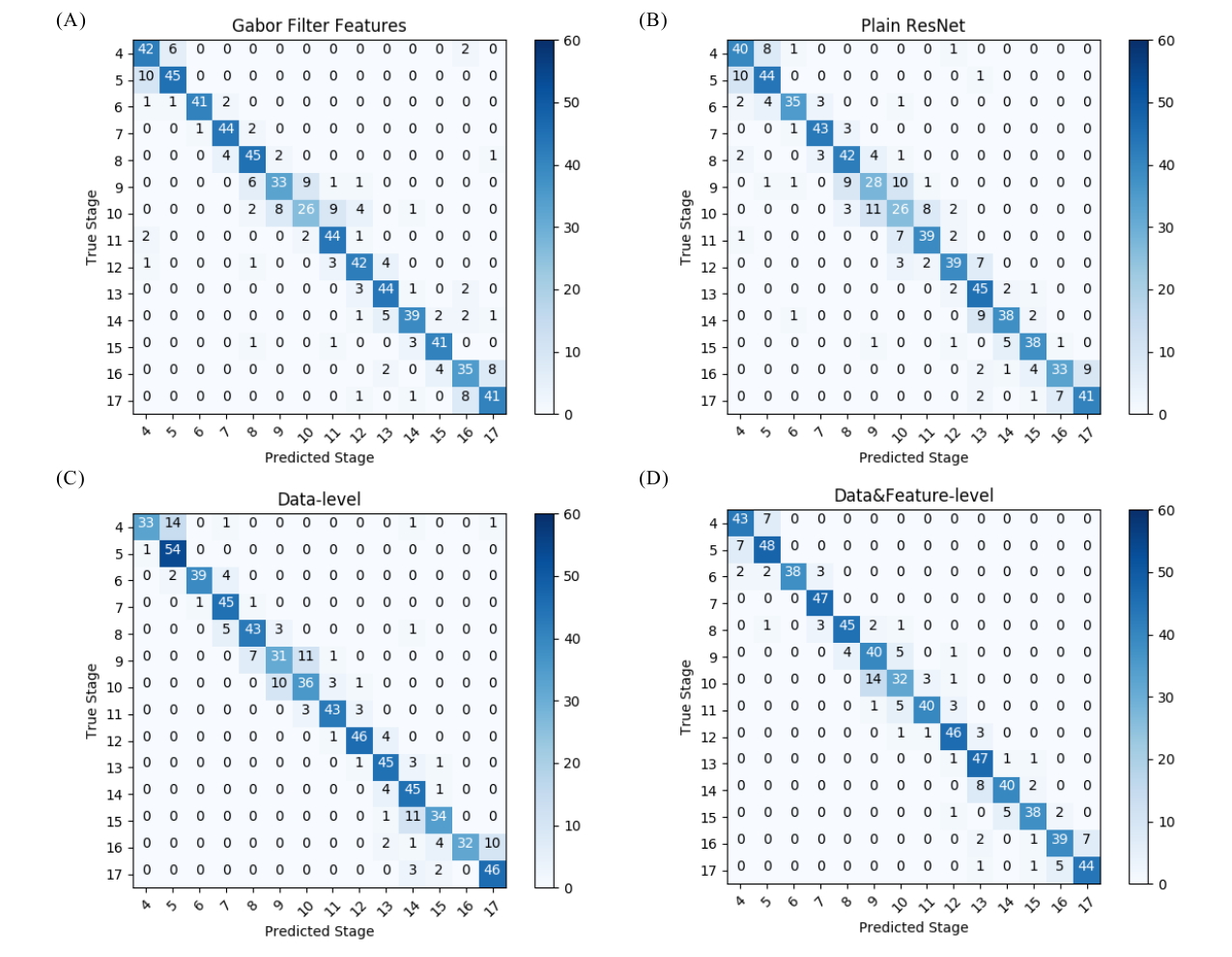}
	\caption{Comparison of performance between different methods in terms of confusion matrices. (A) is our baseline~\cite{yuan2013automated}. (B), (C), and (D) correspond to ResNet models trained using three different approaches: train it directly on $\mathcal{D}_s$~(Plain ResNet), train it using data-level low-shot learning only~(Data-level), and train it using both data-level and feature-level low-shot learning~(Data\&Feature-level).}
	\label{fig: confusion}
\end{figure*}

\begin{figure*}[!ht]
	\centering
	\includegraphics[width=\textwidth]{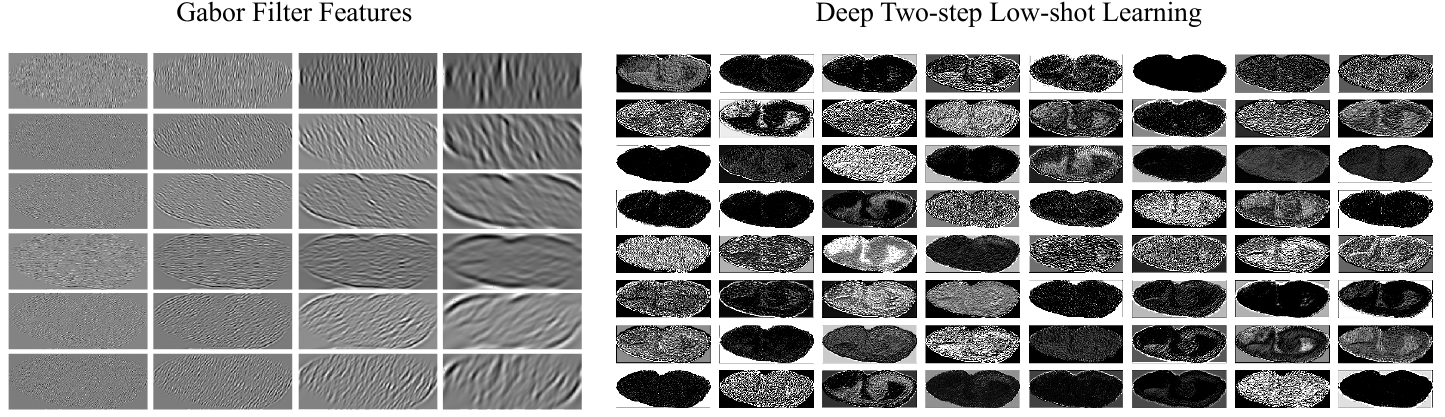}
	\caption{Comparison between feature maps generated by Gabor filters in~\cite{yuan2013automated} and those created by our proposed deep two-step low-shot learning model.}
	\label{fig: feature}
\end{figure*}

\section{Experimental Studies}\label{sec:experiment}

In this section, we evaluate our proposed deep two-step low-shot learning
framework for the precise stage prediction task of ISH images on the Berkeley Drosophila
Genome Project~(BDGP) dataset. We perform qualitative evaluation in terms of the prediction accuracy as well as qualitative evaluation based on feature visualization and genomewide expression maps~(GEMs)~\cite{kumar2011flyexpress,konikoff2012comparison}. Finally, we show the saliency maps and masked GEMs for interpretation and visualization.

\begin{figure*}[t]
	\centering
	\includegraphics[width=\textwidth]{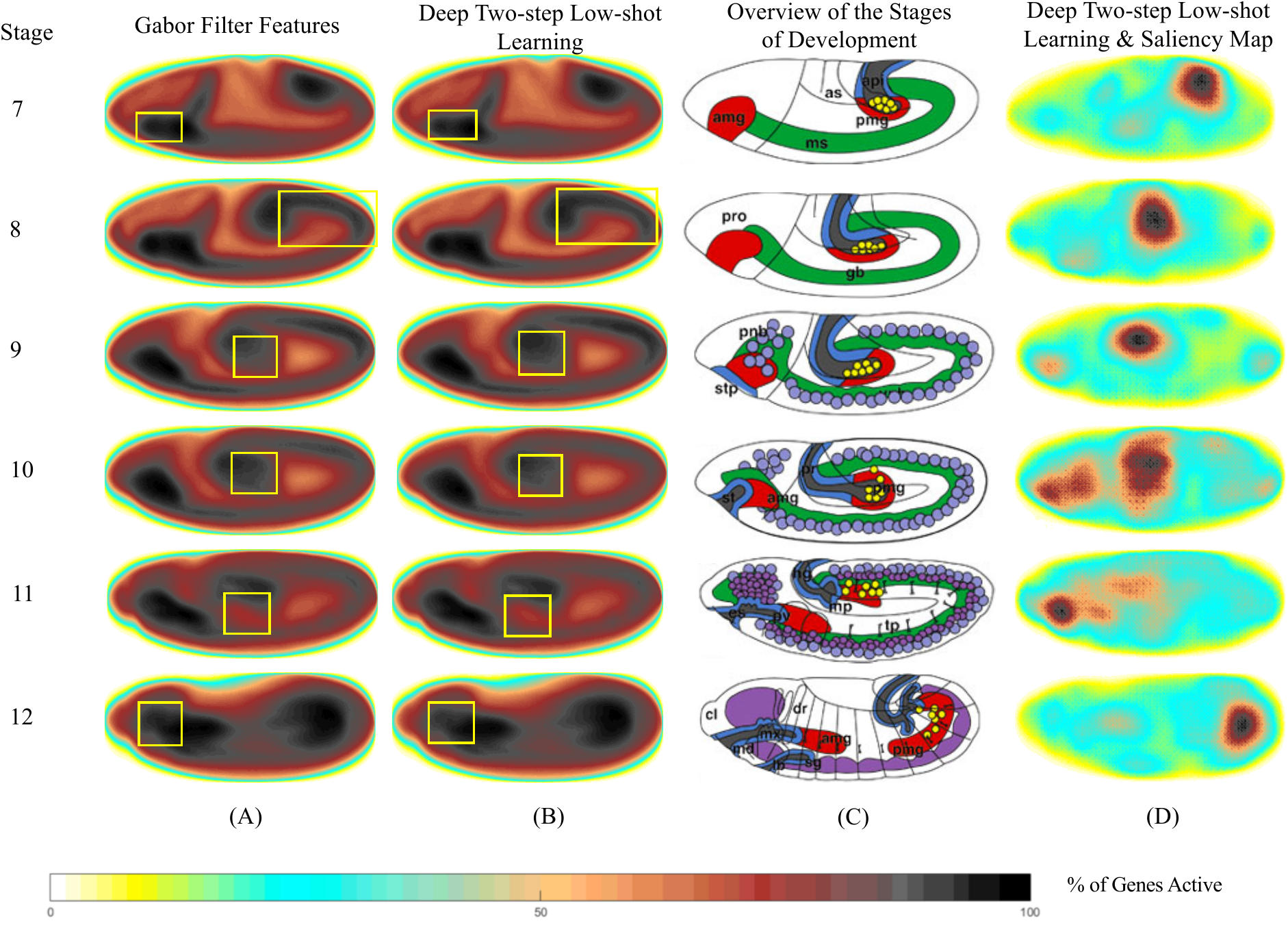}
	\caption{Visualizations of GEMs and masked GEMs generated using different methods for stage 7-12. (A) GEMs generated using prediction results of the baseline. (B) GEMs generated using prediction results of our proposed two-step low-shot learning model. (C) Overview of the stages of development~\cite{hartensteinatlas}. (D) Masked GEMs generated using prediction results of our proposed two-step low-shot learning model and corresponding saliency maps. The differences can better observed when zooming in.}
	\label{fig: s4}
\end{figure*}

\subsection{Experimental Setup}

The BDGP dataset contains 36908 \textit{Drosophila} ISH images in lateral view.
Among them, only 3474 ISH images are annotated with precise stage labels.
In addition, all images are labeled into six stage ranges: 1-3, 4-6, 7-8,
9-10, 11-12 and 13-17 as illustrated in Figure~\ref{fig:embryo}.
According to Section~\ref{sec:application}, we have $\mathcal{D}_t$ with $n_t=36908$ and $\mathcal{D}_s$ with $n_s=3474$, respectively.
The details are listed in Table~\ref{table: data}.
Note that, since the \textit{Drosophila} embryogenesis is not very meaningful from stage 1 to stage 3, the precise stage prediction task does not involving this range. Therefore, there is no precise stage annotation for this range.
All ISH images are standardized by the pipeline proposed in~\cite{konikoff2012comparison}, and the size of processed images is $128 \times 320$.
To perform evaluation, we randomly select $70\%$ of images $\mathcal{D}_s$ for training and test our proposed model on the rest images.

\begin{figure*}[t]
	\centering
	\includegraphics[width=\textwidth]{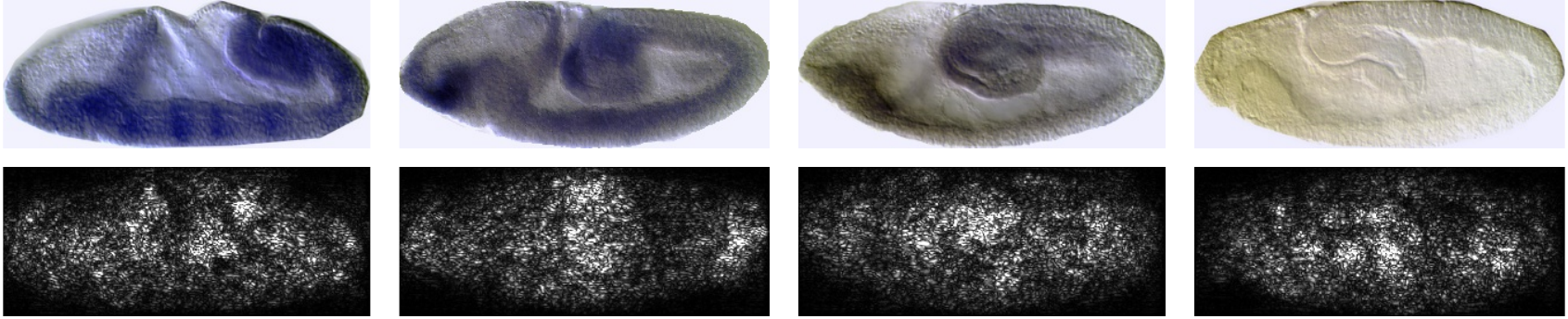}
	\caption{Examples of saliency maps generated by our proposed two-step low-shot learning model. The first row is the original \emph{Drosophila} ISH images. The second row is the corresponding saliency maps. White color indicates that the pixel contributes significantly to the prediction result of the image.}
	\label{fig: salient}
\end{figure*}

We use the previous state-of-the-art model proposed in~\cite{yuan2013automated} that use Gabor filters as our baseline.
For deep learning, we use the ResNet introduced in Section~\ref{sec:application}. To show the effectiveness of our proposed deep two-step low-shot learning approach, we compare the performance of our ResNet trained using three different approaches: train it directly on $\mathcal{D}_s$~(Plain ResNet), train it using data-level low-shot learning only~(Data-level), and train it using both data-level and feature-level low-shot learning~(Data\&Feature-level). For all training processes, we employ the stochastic gradient descent algorithm~\cite{lecun1998gradient}. The learning rate starts with 0.001 and is multiplied by 0.1 at the 20-th and 30-th epoch. The batch size is set to 16.

\subsection{Quantitative Analysis of Predictive Performance}

The precise stage prediction accuracy of each method is reported in Table~\ref{table: res}.
Note that the accuracy of Plain ResNet is lower than that of our baseline model,
proving that deep learning models can hardly achieve satisfactory performance with limited training samples.
Training the ResNet with our data-level deep low-shot learning method addresses this problem effectively and improves the performance significantly. By adding the feature-level learning step, our proposed deep two-step low-shot learning model achieves better predictive performance and becomes the new state-of-the-art model for the precise stage prediction task of ISH images.

In order to analyze the prediction errors of our model, we compare the
confusion matrices of the prediction results, as shown in Figure~\ref{fig: confusion}.
We observe that most wrongly-predicted samples are classified into
neighboring stages, within the correct stage ranges. In particular, for deep learning models trained using our proposed low-shot learning methods, the number of samples that are categorized into wrong stage ranges is fewer than the baseline. It indicates that the data-level deep low-shot learning step effectively incorporates information from samples with stage range labels and improves the performance of precise stage prediction.

\subsection{Qualitative Analysis of Predictive Performance}

We perform two qualitative analyses on the baseline and the proposed deep two-step low-shot learning model.
First, we compare the features extracted by Gabor filters with those generated by the first convolutional layer in the proposed model.
Figure~\ref{fig: feature} provides visualizations of the features. Clearly, Gabor filters can only extract pre-defined low-level texture features while our deep learning model is able to extract data-driven and task-related features automatically. Our deep two-step low-shot learning framework takes advantage of the power of deep learning models by proposing effective training approaches and yields improved performance.

Second, we visualize genomewide expression maps~(GEMs)~\cite{konikoff2012comparison,kumar2011flyexpress} generated by the baseline and our deep two-step low-shot learning model. Computing GEMs relies on the predictive performance and accurate precise stage prediction results produce more biologically meaningful GEMs. The first two columns in Figure~\ref{fig: s4} compare the GEMs generated using prediction results of the baseline and our model. According to the third column, we can see that the developing trend shown in both GEMs is consistent
with that shown in~\cite{hartensteinatlas}. In addition, since our proposed method achieves better predictive performance, the generated GEMs contain more biologically meaningful details as demonstrated by yellow bounding boxes. The differences can be better observed when zooming in the Figure~\ref{fig: s4}.

\subsection{Visualization through Salience Maps and Masked Genomewide Expression Maps}

As introduced in Section~\ref{sec:interpret}, we can compute saliency maps to measure the pixel-wise contributions of an input image to its prediction result.
We visualize some saliency maps generated by our proposed two-step low-shot learning model in Figure~\ref{fig: salient}.
From the saliency maps, we can infer which pixels contribute most to the prediction results.
We propose to apply saliency maps as masks and use these pixels to generate GEMs instead of using the whole input images.
The resulted masked GEMs are shown in the last column in Figure~\ref{fig: s4}.
The masked GEMs also show a consistent developing trend as shown in the third column. Moreover, they provide better visualizations to help identifying and visualizing developmental landmarks.

\section{Conclusion}

In this work, we propose a deep two-step low-shot
learning framework and apply it on the precise stage prediction task of ISH images.
We first propose to employ deep learning models to automatically extract data-driven and task-related features instead of using hand-crafted feature extraction methods. In order to build accurate deep learning models from limited training samples, we formulate the
task as a deep low-shot learning problem.
To solve it, we propose a deep two-step low-shot learning approach that is composed of data-level learning and feature-level learning.
The first data-level learning step leverages extra data to facilitate the training.
The second step, feature-level learning, introduces a regularization method which forces a high similarity of features extracted from different samples in the same class.
The proposed deep two-step low-shot learning framework enables us to effectively train deep learning models with high classification accuracy.
We conduct thorough experiments on the BDGP dataset using a deep residual network as our base model. Both quantitative and qualitative experimental results demonstrate that our proposed deep two-step low-shot model outperforms the previous state-of-the-art model significantly.
In addition, we explore visualization based on saliency maps~\cite{simonyan2013deep}, which compute pixel-wise contributions of input images to the prediction results. In the precise stage prediction task of ISH images, we apply saliency maps to generate masked genomewide expression maps, which provide biologically meaningful visualizations that help identifying and visualizing developmental landmarks ~\cite{wang2016drosophila}.

\section*{Acknowledgment}
This work was supported in part by National Science Foundation
Grants IIS-1908220 and DBI-1661289.

\bibliography{deep}
\bibliographystyle{IEEEtran}

\end{document}